# A Comparative Analysis of Nature-inspired Feature Selection Algorithms in Predicting Student Performance

Thomas Trask
Computer Science Delpt.
Georgia Institute of Technology
Atlanta, GA, USA
thomas.trask@gatech.edu
https://orcid.org/0009-0008-5796-4835

*Abstract*—Predicting student performance is key in leveraging effective pre-failure interventions for at-risk students. As educational data grows larger, more effective means of analyzing student data in a timely manner are needed in order to provide useful predictions and interventions. In this paper, an analysis was conducted to determine the relative performance of a suite of nature-inspired algorithms in the feature-selection portion of ensemble algorithms used to predict student performance. A Swarm Intelligence ML engine (SIMLe) was developed to run this suite in tandem with a series of traditional ML classification algorithms to analyze three student datasets: instance-based clickstream data, hybrid single-course performance with pre-course metadata, and student meta-performance when taking multiple courses simultaneously. These results were then compared to previous predictive algorithms and, for all datasets analyzed, it was found that leveraging an ensemble approach using nature-inspired algorithms for feature selection and traditional ML algorithms for classification significantly increased predictive accuracy while also reducing feature set size by up to 65%. The SIMLe test engine was also able to identify keystone data features useful in creating pre- and intra-course interventions.

## I. INTRODUCTION

Educational instruction exists within a spectrum [1, 2]. On one end of the spectrum is pedagogy, wherein students are led by a teacher down a predefined, controlled path with most of the educational responsibility resting on the teacher with the student being dependent on them for instruction. Due to reduced baseline of individual educational responsibility for students, pedagogy is a common choice for traditional k-12 education. On the other end of the educational spectrum lies heutagogy, in which the student acts as their own teacher, independent and free to chart their own educational path and learn at their own pace. Massive, Open online course (MOOC) programs are useful intra and inter-course example of this paradigm; students are free to explore any topic within a given context and chase any intellectually promising rabbit hole they find. Between pedagogy and heutagogy lies andragogy, where the teacher acts as a facilitator for the student's independent exploration, like apprenticeship-style or university education, wherein the student can leverage both their experiences and the teachers' expertise to fulfill their educational goals. Continually more skillsets are required for students to successfully progress from pedagogical learning onto guided and independent learning. Developing skills in time management, self-guided researching, intrinsic motivation, and manifesting engagement are often underdeveloped in students coming from a fundamentally pedagogical educational framework but are nonetheless critical for a success transition. Facilitated online learning provides a steppingstone beyond the traditional pedagogical paradigm but is often a significant enough step as to cause many students to stumble. Tools are needed to help students effectively bridge this skill gap in order to succeed. Effectively predicting student performance is the first step in helping students fill these meta-educational skill gaps.

Many machine learning frameworks exist within the educational space and are used to predict student success in classes across the Pedagogy-Androgogy-Heutagogy spectrum [3]. In addition, different datasets focus on different aspects. Some datasets focus on individual event-driven data, such as clickstream session data for online classes. Other datasets focus on a student's demographics and their performance on a single course. Still others focus on the student's whole academic career. This project will explore the relative effectiveness of Swarm Intelligence machine learning algorithms on predicting student success within this same scope. Swarm Intelligence (SI) algorithms are a suite of ML optimization/categorization technique used to find a best-case solution using multiple candidate solutions "particles" operating in localized search spaces which iteratively move within their search space based on both their local optimized solution and a global best-known solution, an approach not dissimilar to the flocking of geese in flight. SI algorithms' novelty come from their approach of mimicking living organizational patterns (such as those from bees or fruit flies or bats) to efficiently solve NP-hard problems within the ML space (feature selection, classification). SI algorithms are split into groups that are suitable for single-objective (finding one key feature in a dataset) and multiple-objective classifications (balancing the optimal subset of a series of features). Because of this versatility, SI algorithms have been successfully combined with SI algorithms, non-SI algorithms, and applied to different instances of an algorithm's process to create complex ensemble algorithms and improve performance of existing ensemble algorithms [4, 5, 6].

## II. RELATED WORK

Accurately predicting student performance using machine learning has been analyzed significantly within the literature [7, 8, 9]. Neural networks, deep learning, recommendation engines, and collaborative filtering, etc. have been shown to accurately predict student class performance at 70-80% accuracy [3]. While effective, these 'traditional' algorithms often require large datasets with many features to work [CN]. SI algorithms have been shown to outperform traditional ML algorithms (random forests, decision trees, KNN) generally and particularly in data-sparse applications [1, 5, 10]. This difference in general performance necessitates analyzing SI's relative performance compared to existing research. In addition to being performant on data-sparse datasets, SI algorithms can also be used as part of an ensemble ML approach and have been shown to significantly outperform traditional ML algorithms in large datasets [6, 11, 12, 13, 14, 15]. This necessitates an additional examination between SI algorithms used in ensemble solutions and existing ML ensemble solutions.

III. METHODOLOGY

A. Goals

The below research addresses the question of whether and which Swarm Intelligence algorithms better predict student performance when compared to existing ML frameworks. It seeks to address the following sub-questions in that process: What is the difference in predictive performance between difference Swarm Intelligence Algorithms? How do they perform when compared to existing classification approaches when used against educational data? Are the algorithms able to identify keystone datapoints to build interventions around? If so, which educational/meta-educational factors impact student performance?

B. Datasets

Three separate datasets will be examined, focused on different levels of the learning experience:

instance-based clickstream data measuring how a student interacts with a program, intra-course data covering a student's performance in a single hybrid course, and inter-course data covering a student's performance within a series of courses. These datasets were chosen both for their content availability and, as noted the previous research conducted on them, necessary for a comparative analysis.

C. Testing Methodolgoy

For this project, the SIMLe (Swarm Intelligence ML engine) testing framework was developed. The SIMLe process occurs in 3 distinct phases: pre-processing, feature-selection, and analysis.

During the pre-processing phase, fields with categorical string values are translated into discrete numerical values. Fields with large ranges are binned into discrete values. For instance, individual 0-100 grades may be converted to binary pass/fail (e.g. (0<70 fail, 71>00 pass) or categorical values (90-100 = 5, 80-90 =4, 70-80=3, 60-70 = 2, < 60 = 1

During the feature-selection phase, each nature-inspired algorithm (NIA) studied was executed against the dataset using the NIAPy python package, to determine best-fit feature sets. Then, features with a feature score < 50% were removed from the testing dataset.

During the analysis phase, predictive scores were generated against the feature-limited datasets using the below traditional ML algorithms using the SKLearn python library: K-Nearest Neighbor (KNN), Support Vector Machines (SVM), Decision Trees (DT), Random Forests (RF, neutral, balanced, balanced with subsampling), and Artificial Neural Network (ANN, dual 6-node sigmoid function). Using this predictive performance, it was then determined which NIA/Traditional ML combination worked best for each of the datasets discussed previously.

D. Swarm-Intelligence Alogirthms used

Leveraging the NIAPy python package, the SIMLe process was able to leverage the below Nature-inspired algorithms and were chosen as to cover a wide spectrum of swarm-based algorithm techniques:
- Particle Swarm Optimization (PSO [16]
- Artificial Bee Colony Optimization (ABC) [17]
- Firefly Optimization [18]
- Bat Echolocation Algorithm (BEA) [19]
- Cat Swarm Algorithm (CSA) [20]
- Bacterial Foraging Optimization (BFO) [21]
- Cuckoo Search (Cuckoo) [22]
- Forest Optimization (FOA) [23]
- Monarch Butterfly Optimization (MBO) [24]
- Monkey King Evolution (MKE1) [25]

IV. ANALYSIS

A. MOOC Clickstream Data

XeutangX is China's largest learning platform, hosting almost 3 thousand individual courses with a total course enrollment of 163 **Million** active users. XeutangX hosts a dataset of clickstream data covering 47 million student interactions with the platform, specifically designed to be used to train student dropout prediction models [13].

This dataset was chosen because of its insight into the granularity in which students engage with course content. The dataset analyzed in this project included 2 files. The first file includes a student's individual enrollment information and whether they passed a particular enrollment instance (some students took the same class multiple times). The second file individual timestamped instances of students:
- Navigating the course site
- Loading, starting, pausing, stopping videos,
- loading, passing/failing question prompts,
- engaging with discussion forums.

This instantiated clickstream data was then aggregated to create a per-student total count for each interaction category (32 total). Consolidating the data in this manner resulted in a dataset of **129k enrollment records** for 66k students across **214 courses**.

Feng, et. al's original work used a Context-aware Feature Interaction Network (CFIN), a type of neural network that links information from separate contexts together to predict interactions between them [13]. The researchers linked user's social interaction via class forum activity to determine student friendships. They then used this friendship model to find a high correlation between a student dropping out of a course and whether their friends dropped the course. Using this method, the researchers were able to predict with **86.71%**

| Methods | XuetangX AUC (%) | F1 (%) |
|---|---|---|
| LRC | 82.23 | 89.35 |
| SVM | 82.86 | 89.78 |
| RF | 83.11 | 89.96 |
| DNN | 85.64 | 90.40 |
| GBDT | 85.18 | 90.48 |
| CFIN | 86.40 | 90.92 |
| CFIN-en | **86.71** | **90.95** |

*Figure 1 Original Results on KDDCUP dataset and IPM courses on XeutangX dataset [13]*

Table 1 SIMLe Predicive Performance on Pedagogical Multi-course dataset

| | Features | SVM | RF | DT | KNN | ANN |
|---|---|---|---|---|---|---|
| PSO | 8 | 86.80% | 85.90% | 85.90% | 81.30% | 83.40% |
| ABC | 11 | 83.60% | 84.00% | 83.50% | 79.40% | 80.60% |
| BA | 10 | 86.00% | 85.10% | 84.30% | 80.20% | 82.20% |
| CSO | 1 | 87.10% | 85.90% | 85.50% | 85.20% | 83.40% |
| BFO | 14 | 87.50% | 86.10% | 85.90% | 80.90% | 83.80% |
| Cuckoo | 9 | 88.00% | 87.10% | 86.70% | 82.80% | 84.70% |
| FFA | 9 | 86.80% | 85.50% | 86.00% | 79.90% | 84.00% |
| FOA | 10 | 86.60% | 85.50% | 86.60% | 80.70% | 82.60% |
| MBO | 11 | 88.00% | 87.40% | 86.40% | 83.60% | 84.30% |
| MKE1 | 9 | 85.90% | 85.60% | 81.30% | 81.20% | 82.80% |
| Base | 32 | 85.50% | 86.90% | 85.50% | 79.80% | 82.70% |

accuracy whether a student would drop a MOOC. See Fig. 1 for more results.

Success was measured directly via a field in the first file indicating whether the student dropped the course before completing. Using the NAI-ML ensemble approach discussed above, a predictive **accuracy of 88%** was achieved with a Cukoo search + SVN combination, higher than the implementation accuracy of the original paper. Of note is that the Cat Swarm Optimization/SVN ensemble achieved a predictive accuracy of **87.1%,** higher than the original paper, while using only one feature (total count of site access). See Table 1 for complete results.

*B. Heutagological Hybrid Class Pass/Fail Prediction*

This dataset includes 667 student records split across two classes (Maths, 442 records and Portuguese, 223 records), encompassing both online and on-campus performance. In addition to 2 intermediate grades and one final grade (measured in 0-20), this dataset includes a series of student meta-heuristics (age, sex, urban/rural address, family size, whether the class was paid for or free, hours spent study-ing, whether the student had outside activities, etc.). [14]

In their analysis, the researchers generated 3 data subsets, set A where all grades were present, set B where G2 was missing, and set C where both G1 and G2 were messing. They then analyzed the course data using naïve prediction (NV), Neural Networks (NN), SVM, Decision Trees (DT), Random Forests (RF), and. They achieved a pass/fail predictive accuracy of 70-91%. See Fig. 2 for complete results.

Using a **Cuckoo Search** feature selection and **Decision Tree** predictive model, achieved an impressive 99.2% pass/fail accuracy, a significant improvement over the original research. Additionally, using either a combination of either **Cuckoo Search**/**Random Forests** or **Bat**

**Algorithm/Random Forests** resulted in a similar predictive accuracy of 98.5%. See Table 2 for complete results.

| Input Setup | Mathematics | | | | |
|---|---|---|---|---|---|
| | NV | NN | SVM | DT | RF |
| A | $91.9^{\dagger}_{\pm 0.0}$ | $88.3_{\pm 0.7}$ | $86.3_{\pm 0.6}$ | $90.7_{\pm 0.3}$ | $91.2_{\pm 0.2}$ |
| B | $83.8^{\dagger}_{\pm 0.0}$ | $81.3_{\pm 0.5}$ | $80.5_{\pm 0.5}$ | $83.1_{\pm 0.5}$ | $83.0_{\pm 0.4}$ |
| C | $67.1_{\pm 0.0}$ | $66.3_{\pm 1.0}$ | $70.6^{*}_{\pm 0.4}$ | $65.3_{\pm 0.8}$ | $70.5_{\pm 0.5}$ |
| Input Setup | Portuguese | | | | |
| | NV | NN | SVM | DT | RF |
| A | $89.7_{\pm 0.0}$ | $90.7_{\pm 0.5}$ | $91.4_{\pm 0.2}$ | $93.0^{\dagger}_{\pm 0.3}$ | $92.6_{\pm 0.1}$ |
| B | $87.5_{\pm 0.0}$ | $87.6_{\pm 0.4}$ | $88.0_{\pm 0.3}$ | $88.4_{\pm 0.3}$ | $90.1^{\dagger}_{\pm 0.2}$ |
| C | $84.6_{\pm 0.0}$ | $83.4_{\pm 0.5}$ | $84.8_{\pm 0.3}$ | $84.4_{\pm 0.4}$ | $85.0^{*}_{\pm 0.2}$ |

Fig. 2 Previous Hybrid Class Predictive performance, [14]

*C. Pedagogical Multi-course Performance Prediction*

This dataset was chosen because it should allow us to gain insight on how taking courses simultaneously impacts student performance [15]. The dataset contains individual student records including:

- Grade information for 5 courses (Physics, Maths, Chemistry, Biology, Intro to Engineering, each consisting of 2 intermediate grades and 1 final grade based on a 100-point scale.
- Answers to the survey questions in Appendix A

To analyze the data, the original researchers used an Optimum Multilabel Ensemble Model wherein a dataset is split into a series of training models, each model is processed individually, and then combined in an ensemble method (majority vote). The re-combined dataset is then evaluated for performance using the base-level classifiers [ label-power sets (LP), binary relevance (BR), classifier chains (CC), and the improved LP (LP+ RAkELo)] in tandem with common ML algorithms [Random Forests (RF), SVM, K-nearest Neighbor (KNN) and Multi-layer Perceptrons (MLN), a type of neural network], The researchers achieved a **79% F1 score using the improved LP/SVM ensemble method**. See Fig. 3 for complete results.

To determine SIMLe's relative performance, a binary success criteria was used with a bifurcation at 50% (>50% passed, <=50% fail). While more traditional ML algorithms failed to generate a better predictive model than the one used in the original research, **SIMLe ensemble methods using**

Table 2 SIMLe Predicive Performance on Heutagogical Hybrid Course Dataset

| | Features | SVM | RF | DT | KNN |
|---|---|---|---|---|---|
| PSO | 15 | 90.8% | 96.9% | 93.1% | 87.7% |
| ABC | 16 | 91.5% | 97.7% | 95.4% | 89.2% |
| BA | 19 | 92.3% | **98.5%** | 95.4% | 90.8% |
| CSO | 7 | 96.2% | 96.2% | 96.2% | 91.5% |
| BFO | 18 | 90.8% | 95.4% | 93.1% | 84.6% |
| Cuckoo | 11 | 91.5% | **98.5%** | **99.2%** | 90.8% |
| FFA | 11 | 93.8% | 96.9% | 95.4% | 90.8% |
| FOA | 8 | 93.8% | 98.5% | 95.4% | 95.4% |
| MBO | 16 | 93.8% | 98.5% | 95.4% | 87.7% |
| MKE1 | 11 | 93.8% | 95.4% | 93.8% | 93.8% |
| Base | 18 | 94.6% | 96.9% | 96.9% | 90.0% |

Table 3 SIMLe Predicive Performance on Pedagogical Multi-course dataset

| | Features | SVM | RF | DT | KNN | ANN |
|---|---|---|---|---|---|---|
| PSO | 4 | 63.0% | 62.2% | 62.2% | 60.1% | **95.1%** |
| ABC | 3 | 62.3% | 65.0% | 64.3% | 28.0% | **95.8%** |
| BA | 6 | 62.2% | 62.2% | 57.3% | 63.6% | 3.5% |
| CSO | 3 | 61.5% | 64.3% | 64.3% | 65.0% | **95.8%** |
| BFO | 4 | 66.4% | 67.8% | 66.4% | 67.1% | **98.6%** |
| Cuckoo | 4 | 70.6% | 65.0% | 65.0% | 63.6% | **98.6%** |
| FFA | 6 | 68.6% | 67.8% | 62.2% | 67.1% | **97.2%** |
| FOA | 5 | 58.1% | 61.5% | 58.0% | 63.6% | **96.5%** |
| MBO | 4 | 65.7% | 63.6% | 63.6% | 60.1% | **95.8%** |
| MKE1 | 6 | 65.8% | 58.0% | 53.1% | 60.1% | **98.6%** |
| Base | All | 68.5% | 67.1% | 53.8% | 67.8% | **97.2%** |

| Problem transformation | Base classifier | Evaluation measures | | | | |
|---|---|---|---|---|---|---|
| | | Hamming | Jaccard | Accuracy | $F1_{micro}$ | $F1_{macro}$ |
| LP | RF | 0.324 ± 0.040 | 0.619 ± 0.061 | 0.231 ± 0.047 | 0.776 ± 0.042 | 0.754 ± 0.052 |
| | SVM | **0.320 ± 0.070** | **0.649 ± 0.081** | 0.264 ± 0.080 | **0.787 ± 0.061** | **0.765 ± 0.072** |
| | KNN | 0.334 ± 0.027 | 0.603 ± 0.046 | 0.186 ± 0.034 | 0.760 ± 0.034 | 0.727 ± 0.047 |
| | MLP | 0.334 ± 0.033 | 0.590 ± 0.063 | 0.220 ± 0.049 | 0.759 ± 0.041 | 0.735 ± 0.050 |
| BR | RF | 0.314 ± 0.029 | 0.620 ± 0.052 | **0.206 ± 0.027** | 0.775 ± 0.036 | 0.745 ± 0.045 |
| | SVM | **0.307 ± 0.041** | **0.638 ± 0.059** | 0.203 ± 0.033 | **0.788 ± 0.041** | **0.760 ± 0.048** |
| | KNN | 0.340 ± 0.035 | 0.582 ± 0.057 | 0.188 ± 0.033 | 0.749 ± 0.042 | 0.723 ± 0.051 |
| | MLP | 0.336 ± 0.034 | 0.589 ± 0.053 | 0.196 ± 0.034 | 0.756 ± 0.038 | 0.726 ± 0.045 |
| CC | RF | 0.313 ± 0.031 | 0.618 ± 0.052 | 0.228 ± 0.030 | 0.777 ± 0.036 | 0.750 ± 0.045 |
| | SVM | **0.307 ± 0.040** | **0.640 ± 0.055** | **0.236 ± 0.035** | **0.789 ± 0.039** | **0.765 ± 0.047** |
| | KNN | 0.335 ± 0.027 | 0.585 ± 0.054 | 0.189 ± 0.023 | 0.754 ± 0.037 | 0.729 ± 0.048 |
| | MLP | 0.336 ± 0.030 | 0.585 ± 0.054 | 0.199 ± 0.036 | 0.755 ± 0.038 | 0.727 ± 0.046 |
| LP + RAkEL$_o$ | RF | 0.314 ± 0.063 | 0.652 ± 0.078 | 0.254 ± 0.069 | 0.792 ± 0.056 | 0.771 ± 0.067 |
| | SVM | **0.312 ± 0.064** | **0.654 ± 0.079** | 0.250 ± 0.072 | **0.794 ± 0.056** | **0.774 ± 0.066** |
| | KNN | 0.328 ± 0.037 | 0.616 ± 0.058 | 0.196 ± 0.038 | 0.769 ± 0.042 | 0.738 ± 0.056 |
| | MLP | 0.321 ± 0.056 | 0.640 ± 0.074 | 0.242 ± 0.062 | 0.784 ± 0.054 | 0.761 ± 0.065 |

Fig 3 Previous Ensemble Method Multi-class Predictive performance (Yuki et al, 2021)

**ANNs achieved a >95% accuracy**, with a Cuckoo Search/ANN combination achieving a 98% accuracy. See Table 3 for complete results.

## V. CONCLUSION

The above results strongly indicate that there is an incentive to integrate nature-inspired algorithms into existing dropout prediction pipelines. The ensemble method implemented in SIMLe outperformed the best results of all 3 of the originally cited papers while being less complicated to implement, and easier to extend. With an average feature-set reduction of 2/3, the SIMLe ensemble algorithms effectively executed on both robust and feature-sparser datasets. Cukoo Search generated the highest predictions the most often and often had the lowest feature set size.

The secondary goal of the research was to identify individual intra/meta-course datapoints that may help develop pre-course interventions. In addition to the predictive performance improvements offered by SIMLe, it can rapidly highlight the predictive nature of specific features, which will aid in developing pre- and intra-course interventions. For instance, in the multi-class androgogical dataset, it showed a stronger correlation between a student's extra-curricular activities and performance than on student's finances or whether they commute. This would indicate that developing a pre-course intervention in effective time management may be useful. In the Hybrid pedagogical course, it was found that a student's gender was a strong indicator of their performance, which may point to intra-course friction driving down female performance or pre-course systemic under-preparation, both of which would need to be addressed on a meta-course level.

## VI. FUTURE WORK

Further work should include analyzing the feature-set weight maps that were generated by the SIMLe test harness to determine what type of data each algorithm responds to (low/high-dimensional categorical, binary, etc). Using the timestamp element in the clickstream data can be used to determine how long a user is active on the platform, which may correlate to student success. The XeutangX dataset tracks users across multiple courses. Additional NAIs and traditional ML classification algorithms should also be implemented. Analyzing the student's progress through the courses may provide useful information on successful course progressions. Executing this ensemble suite against district-wide student datasets may yield insights into important academic trends. Runtime analysis should be done to see whether the developed ensemble approach is faster than those outlined in the research. Feature sets were chosen based on a feature candidacy>0.5. Additional exploration should be performed to determine if that can be increased without reducing accuracy.

The SIMLe test harness proved to be an invaluable tool for this analysis. Future work should focus on disseminating the tool for wider use and developing key features for it such as gui-based analysis tools and feature highlighting.

## VII. ACKNOWLEDGEMENTS

Thomas Trask would like to thank Sheila Foley and Dr. David Joyner for their support and guidance during the development of this project.

## VIII. APPENDIX

### A. Survey Questions for Multi-Course Performance prediction dataset

| Feature | Type | Possible values |
|---|---|---|
| Gender | Categorical (nominal) | Male or Female |
| Age | Numerical | Range: [1, 100] |
| Courses | Numerical | [0, 100] |
| Quality of education | Categorical (nominal) | {Excellent, Very good, Good, Satisfactory, Bad, Very bad} |
| Legal guardians | Categorical (nominal) | {mother and father, father only, mother only, siblings, other, live alone} |
| Family income | Categorical (ordinal) | {<5000, 5000-10000, 10000-20000, >20000} |
| Family educational background | Categorical (ordinal) | {diploma, degree, masters, PhD, high school, high school dropout, no education} |
| Tutorial | Categorical (nominal) | {Yes, No} |
| Grade 10 GPA | Categorical (ordinal) | {2-2.5, 2.5-3, 3-3.5, 3.5-4} |
| Parent occupation | Categorical (nominal) | {Civil servant, Artisan, Trading/merchant, Military} |
| Student's perception towards education | Categorical (nominal) | {Yes, No} |